\documentclass[journal]{IEEEtran}
\usepackage{cite}

\ifCLASSINFOpdf
   \usepackage[pdftex]{graphicx}
\else
\fi
\usepackage{amsmath}
\usepackage{amsfonts}
\usepackage{setspace}

\usepackage{soul} 
\usepackage{booktabs} 

\usepackage[colorlinks=true,linkcolor=blue,citecolor=blue,urlcolor=black ]{hyperref}

\usepackage{graphicx}

\usepackage{subcaption}
\usepackage{epsfig}
\usepackage{calc}
\usepackage{amssymb}
\usepackage{amstext}
\usepackage{amsthm}
\usepackage{multicol}
\usepackage{multirow}
\usepackage{pslatex}
\usepackage{siunitx}

\begin{document}
%
\title{Indoor Point-to-Point Navigation with Deep Reinforcement Learning and Ultra-wideband}
%
%
%

\author{Enrico~Sutera, Vittorio~Mazzia, Francesco~Salvetti, Giovanni~Fantin
        and~Marcello~Chiaberge
\thanks{The authors are with Politecnico di Torino -- Department of Electronics and Telecommunications, PIC4SeR, Politecnico di Torino Interdepartmental Centre for Service Robotics and SmartData@PoliTo, Big Data and Data Science Laboratory, Italy. Email: \{name.surname\}@polito.it.}}

%
%

\markboth{}%
{Enrico Sutera \MakeLowercase{\textit{et al.}}: Indoor Point-to-Point Navigation with Deep Reinforcement Learning and Ultra-wideband}
%



\maketitle

\begin{abstract}
Indoor autonomous navigation requires a precise and accurate localization system able to guide robots through cluttered, unstructured and dynamic environments. Ultra-wideband (UWB) technology, as an indoor positioning system, offers precise localization and tracking, but moving obstacles and non-line-of-sight occurrences can generate noisy and unreliable signals. That, combined with sensors noise, unmodeled dynamics and environment changes can result in a failure of the guidance algorithm of the robot. We demonstrate how a power-efficient and low computational cost point-to-point local planner, learnt with deep reinforcement learning (RL), combined with UWB localization technology can constitute a robust and resilient to noise short-range guidance system complete solution. We trained the RL agent on a simulated environment that encapsulates the robot dynamics and task constraints and then, we tested the learnt point-to-point navigation policies in a real setting with more than two-hundred experimental evaluations using UWB localization. Our results show that the computational efficient end-to-end policy learnt in plain simulation, that directly maps low-range sensors signals to robot controls, deployed in combination with ultra-wideband noisy localization in a real environment, can provide a robust, scalable and at-the-edge low-cost navigation system solution.
\end{abstract}

\begin{IEEEkeywords}
Indoor Autonomous Navigation, Autonomous Agents, Deep Reinforcement Learning, Ultra-wideband
\end{IEEEkeywords}

%

\section{Introduction}
\label{sec:introduction}
The main focus of service robotics is to assist human beings, generally performing dull, repetitive or dangerous tasks, as well as household chores. In most of the applications, the robot has to navigate in an unstructured and dynamic environment, thus requiring robust and scalable navigation systems. In practical application, robot motion planning in dynamic environments with moving obstacles adopts a layered navigation architecture where each block attempts to solve a particular task. In a typical stack, in a GPS-denied scenario, precise indoor localization is always a challenging objective with a great influence on the overall system and correct navigation \cite{rigelsford2004introduction}. Indeed, algorithms such as SLAM \cite{cadena2016past} or principal indoor localization techniques based on technologies, such as WiFi, radio frequency identification device (RFID), ultra-wideband (UWB) and Bluetooth \cite{zafari2019survey}, are greatly affected by multiple factors; among others, presence of multi-path effects, noise and characteristics of the specific indoor environment are still open challenges that can compromise the entire navigation stack. 

Robot motion planning in dynamic and unstructured environments with moving obstacles has been studied extensively \cite{mohanan2018survey}, but, being an NP-complete \cite{barraquand1991robot} problem, classical solutions have significant limitations in terms of computational request, power efficiency and robustness at different scenarios. Moreover, currently available local navigation systems have to be tuned for each new robot and environment \cite{chen2015deepdriving} constituting a real challenge in presence of dynamical and unstructured environments.

Deep learning and in particular Deep reinforcement learning (RL) has shown very promising results in fields as diverse as video games \cite{mnih2015human,mnih2013playing}, energy usage optimization \cite{mocanu2018line}, remote sensing \cite{salvetti2020multi,khaliq2019refining,mazzia2020improvement} and visual navigation \cite{zhu2017target,tamar2016value,aghi2020local}, since 2013. Greatly inspired by the work of Chiang et al. \cite{chiang2019learning}, we exploited deep reinforcement learning to obtain an agent robust to localization noise and able to map raw noisy low-level 2-D lidar observations to robot controls linear and angular velocities. Indeed, the obtained learnt policy through a plain and fast simulation process is a light-weight, power-efficient motion planning system that can be deployed at the edge, on very low-cost hardware with limited computational capabilities. 

In particular, we focused our research on a tight integration between the point-to-point local motion planner, learnt in simulation, with UWB localization technology, providing experimental proofs of the feasibility of the system UWB-RL in a real setting. UWB radios are rapidly growing in popularity, offering decimeter-level accuracy and increasingly smaller and cheaper transceivers \cite{magnago2019robot}. In comparison with other techniques, UWB enables both distance estimation and communication among devices within the same radio chip with relative low-level consumption. However, the accurate estimation of the position of a robot is critical for its correct navigation and, as previously mentioned, also UWB, in a real scenario, is affected by several factors of disturbance.
Our results show that, even in the presence of very uncertain localization information, due to the presence of moving obstacles in the environment, multi-path effects and other sources of noise, our proposed solution is robust and has comparable performance with classical approaches. Nevertheless, our solution has a much lower computational request and power consumption constituting a competitive and end-to-end local motion planner solution for indoor autonomous navigation in dynamic and unknown environments.



\section{Proposed Method}
\label{section:sec3}
\subsection{Reinforcement Learning}
\begin{figure}
    \centering
    \includegraphics[width=1\linewidth]{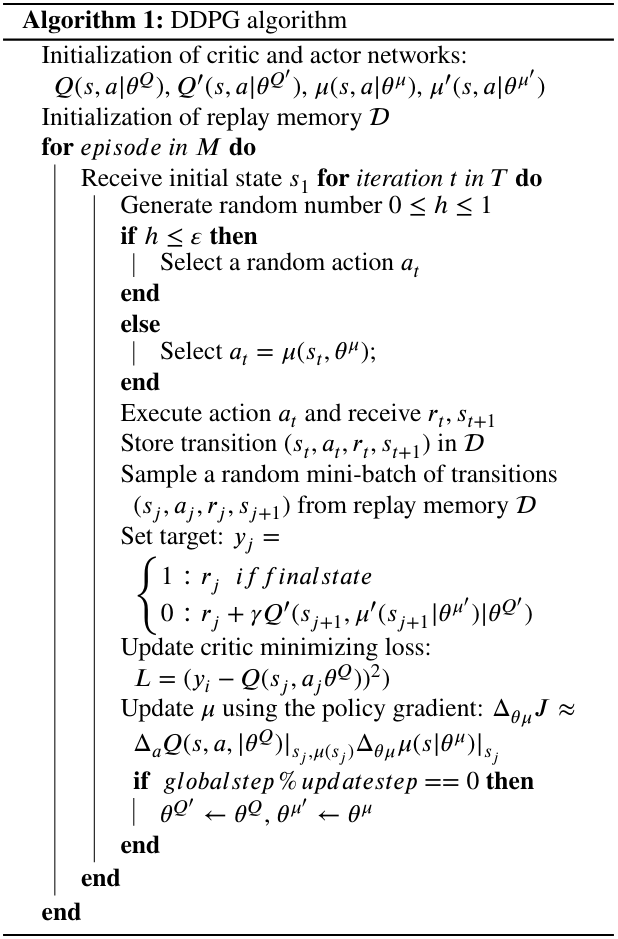}
\end{figure}
Deep RL is a machine learning technique that merges deep learning and reinforcement learning together. The latter is generally used for tackling problems that can be modeled as a Markov decision process (MDP). Hence, the typical learning setup consists of an agent which interacts with an environment. The agent selects an action $a_t \in \mathcal{A}$ and performs it in the environment, which gives back a new state $s_{t+1} \in \mathcal{S}$ and a reward $r_{t+1}$, sequentially at each time-step $t$. The environment may also be stochastic. $\mathcal{A}$ and $\mathcal{S}$ are the space of the actions and the space of the states, respectively. The reward $r_t$ is the feedback signal at the basis of the learning process from raw data, hence being higher for "better" actions and lower for "worse" ones. The agent chooses an action by following a policy $\pi$ that maps states to actions. A sequence with shape $s_0,a_0,r_1,s_1,a_1,\dots, s_i, a_i, r_{i+1}, s_{i+1}$ is then generated, which can be seen as many transitions one after another. The training phase aims at making the agent learn to maximize the return $G$, which is usually the discounted sum of future rewards, expressed as
\begin{equation}
    G = \sum_{k=0} \gamma^k R_{t+k+1}
\end{equation}

The term $\gamma$ is called discount factor and it regulates the importance of rewards along the episode. It can assume values between $0$ (only the immediate reward is important) and $1$ (all future rewards are equally important).
The agent is typically characterized by a policy $\pi(a|s)$ which maps states to actions. A policy can be stochastic, e.g. can give the probability of an action $a$ to be taken when in state $s$, or deterministic, hence giving the action directly and in this case is often denoted by $\mu$.

Since the aim of an agent is to maximise $G$, it is useful to define the expected return when an action $a_t$ is taken from a state $s_t$ and then a policy $\pi$ is followed. This is expressed by the action-value function:

\begin{equation}
    Q^\pi(s_t,a_t) = \mathbb{E}_{r_{i\geq t}, s_{i>t} \sim E, a_{i>t}\sim\pi} [ R_t \mid s_t,a_t]
\end{equation}

In many RL approaches the Bellman equation is used:

\begin{equation}
\begin{split}
     Q^\pi(s_t,a_t) = \; &\mathbb{E}_{r_t, s_{t+1} \sim E}[r(s_t,a_t) \\
     & + \gamma \mathbb{E}_{a_{t+1}\sim \pi}[Q^\pi(s_{t+1},a_{t+1})]]
\end{split}
\end{equation}

\noindent which, under target deterministic policy becomes:

\begin{equation}
\begin{split}
    Q^\mu(s_t,a_t) = \; & \mathbb{E}_{r_t, s_{t+1} \sim E}[r(s_t,a_t) \\
    & + \gamma [Q^\mu(s_{t+1},a_{t+1})]]
\end{split}
\end{equation}

This relationship is used to learn $Q^\mu$ off policy, that means that the exploited transition can also be generated by using another stochastic behavioural policy $\beta$. This approach can be referred to as Q-learning. Considering a finite action space, once the Q function is known, it is sufficient to choose the action that maximizes the expected returns. This is also called greedy policy:

\begin{equation}
\mu(s)=arg {max}_a Q(s,a)
\end{equation}

If we consider to approximate the action-value function using a function approximator, whose parameters can be denoted as $\theta^Q$, the optimization can be performed by minimizing the loss, which can be expressed as:

\begin{equation}
    L(\theta^Q) = \mathbb{E}_{s_t \sim \rho^\beta, a_t \sim \beta, r_t \sim E}[(Q(s_t,a_t \mid \theta^Q)-y_t)^2]
\end{equation}

\noindent where:

\begin{equation}
y_t = r(s_t, a_t)+ \gamma Q(s_{t+1}, \mu(s_{t+1}\mid\theta^Q))
\end{equation}

\noindent and $\rho$ denotes the discounted state visitation distribution for a policy $\beta$.
This procedure was recently used along with two new feature: a \textit{replay buffer} and a \textit{target network} for obtaining the target $y_t$ \cite{mnih2013playing}\cite{mnih2015human}.

\begin{figure*}
    \centerline{\includegraphics[scale=0.35]{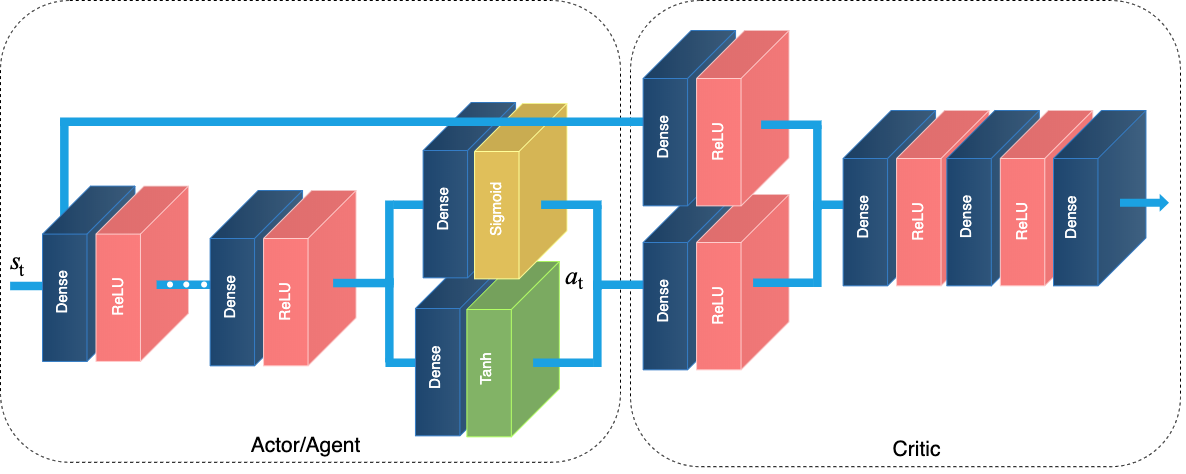}}
    \caption{Graphical representation of the actor/critic architecture. During the training procedure, the actor processes the observations of the robot $s_{t}$ with a cascade of fully connected layers producing distinct actions $a_{t}$ for the angular and linear velocity. Subsequently, the critic network takes as input both $s_{t}$ and $a_{t}$ generating the corresponding Q value estimation. After the training procedure, the policy learnt by the agent in simulation is exploited by the robot to navigate from point-to-point. }
    \label{fig:actor-critic architecture}
\end{figure*}

\begin{figure}
    \centering
    \includegraphics[width=0.75\linewidth]{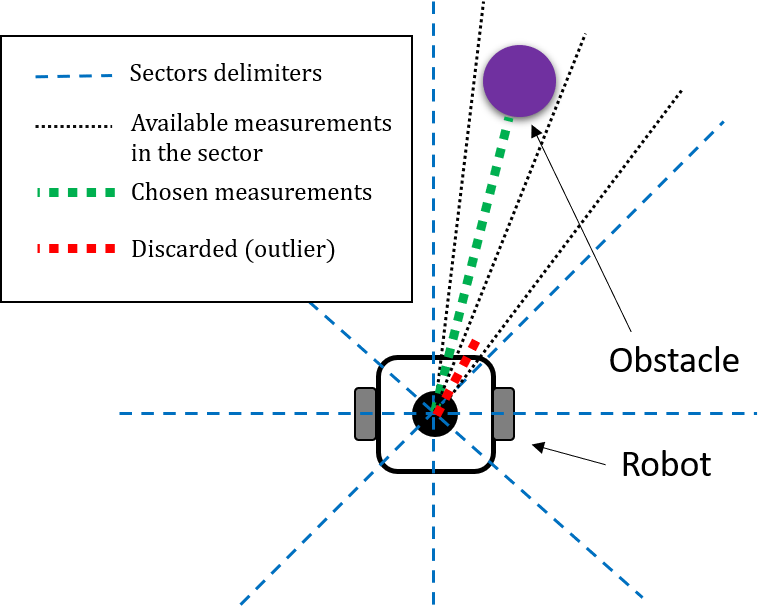}
    \caption{Scheme of used lidar measurements. Lower values of distance are considered more significant for obstacle detection.}
    \label{fig:lidar_min}
\end{figure}

\subsection{Deep Deterministic Policy Gradient}
The above seen Q-learning-related procedure cannot be directly applied to a problem with a continuous action space.
So, we implement a version of the deep deterministic policy gradient (DDPG) algorithm \cite{DDPG} that uses an actor-critic approach to overcome the limitations of discrete actions. Considering to be using function approximators, actor and critic can be denoted respectively as $ \mu(s\mid\theta^\mu)$ and $Q(s, a \mid \theta^Q)$.
The critic function is learned as done in Q-learning, hence using the Bellman equation and exploiting the same loss. The actor function instead is updated by exploiting the knowledge of the policy gradient \cite{SilverDDPG}. Considering a starting distribution $J=\mathbb{E}_{r_i,s_i\sim E, a_i \sim \pi}[R_1]$ and applying the chain rule to the expected return with respect to the parameters of the actor, the policy gradient can be obtained:

\begin{equation}
\begin{split}
\Delta_{\theta\mu}J &\approx \mathbb{E}_{s_t \sim \rho^\beta}[\Delta_{\theta\mu}Q(s,a\mid\theta^Q)\mid_{s=s_t, a=\mu(s_t\mid\theta^\mu)}]\\
    &= \mathbb{E}_{s_t \sim \rho^\beta}[\Delta_a Q(s,a\mid\theta^Q)\mid_{s=s_t, a=\mu(s_t)} \\
      & \,\quad \cdot \Delta_{\theta_\mu}\mu(s\mid\theta^\mu)\mid_{s=s_t}]
\end{split}
\end{equation}

The full pseudo-code is shown in Alg. 1.
We hardly update the target networks periodically instead of performing a continuous soft update. Moreover, we tackle the exploration-exploitation dilemma by maintaining an  $\varepsilon$ probability to perform a random action rather than following the policy $\mu$. The value of epsilon decays during the training as:

\begin{equation}
\varepsilon = \max(\varepsilon_0 \varepsilon_{d}^{episode}, \varepsilon_{min})
\end{equation}

\noindent where $\varepsilon_d$ is the decay parameter.

\subsection{Point-to-Point Agent Training}
In this work, we deal with continuous domains, holding $ \mathcal{A} \in \mathbf{R}^N$ (continuous control) and $ \mathcal{S} \in \mathbf{R}^M$ (continuous state space), with $N$ and $M$ dimensions of action and observation spaces. Concerning the latter, the observation is the representation that the agent has of the current state. In our case, the observation is a vector with 62 elements. It is made of 60 1-D measurements of the lidar, and distance and angle with respect to the goal. During the simulation phase, we use odometry data and magnetometer measurements to compute and provide the previously mentioned distance and angle. This clearly demonstrates the robustness of the trained agent, which is able to generalize to the real scenario even without explicitly modelling UWB localization signals during the training process. The 1-D 60 measurements are not equally spaced. Instead, the whole $2\pi$ circle is split into 60 sectors, and the minimum non-outliers are taken, to guarantee the knowledge of nearer obstacles, as shown in Fig. \ref{fig:lidar_min}.\\

The action instead is a 2-D vector, containing the angular and the linear velocity of the robot:
\begin{itemize}
    \item Linear velocity: the \textit{sigmoid} activation function guarantees a value between $0$ and $1$ since we want the robot to only have non-negative values of speed;
    \item Angular velocity: the \textit{hyperbolic tangent} activation function constraints the output between $-1$ and $1$.
\end{itemize}
According to the \textit{target network} technique, we use four networks: actor network, critic network and their target twins, with the same architectures represented in Fig. \ref{fig:actor-critic architecture}.
The networks are mainly constituted by fully connected layers with ReLU activation functions, except for the final ones. The first three layers of the actor have respectively 512, 256 and 256 neurons. The critic first two hidden layers have respectively 256 (state side) and 64 (action side) units. The following hidden layers have 256 and 128 neurons sequentially. The last layer of the critic is a single output FC with linear activation function to provide the $Q$ value.

\begin{figure}
    \centering
    \includegraphics[width=0.7\linewidth]{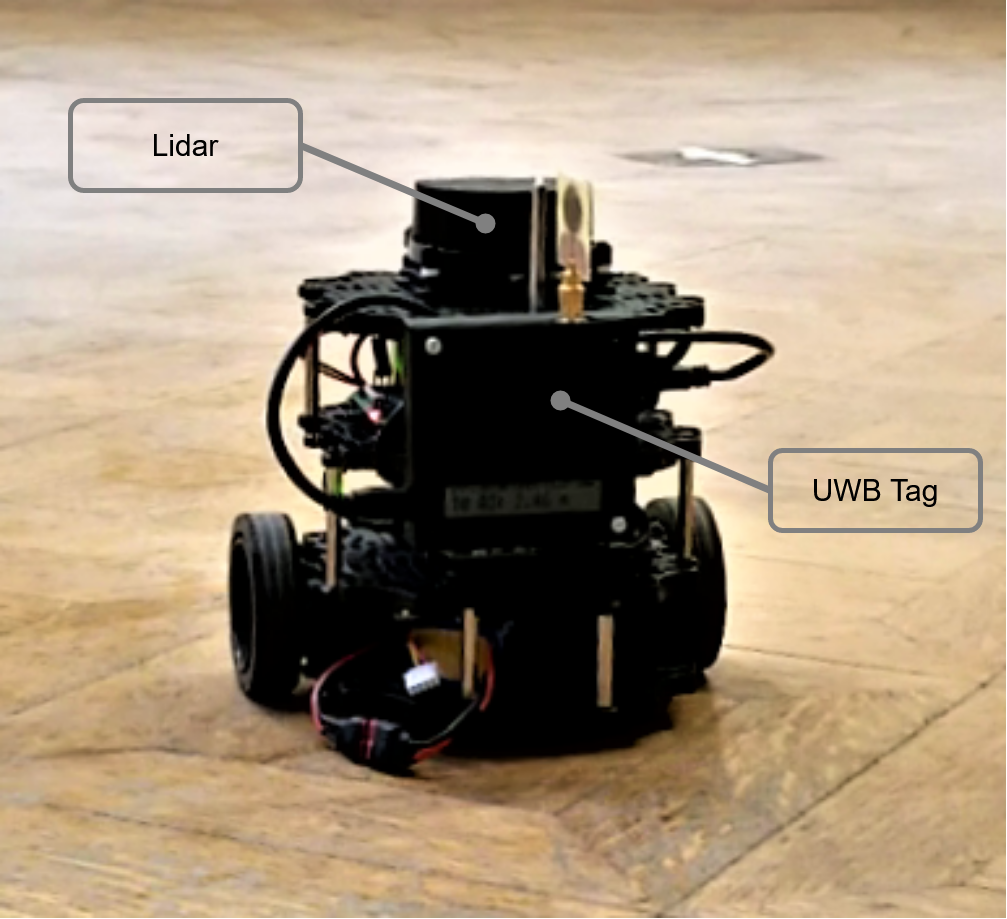}
    \caption{The robotic platform used for the experimentation: a Robotis TurtleBot3 Burger with a Decawave EVB1000 Ultra-wideband tag.}
    \label{fig:robot}
\end{figure}

\section{Experimental Discussion and Results}
In this section, we present the hardware and software setup used during the experimentation phase. We provide a full description of the training phase of the RL agent, with a detailed list of all the selected hyperparameters. Finally, we describe the different tests performed and we present a quantitative evaluation of the proposed local planner.
\label{section:sec4}

\subsection{Hardware and robotic platform}

The training of the RL agent is performed using a workstation with an Intel Core i7 9700k CPU, along with 64 GB of RAM. It takes around 24 hours to complete. Concerning the robotic platform, we select the Robotis TurtleBot3 Burger model\footnote{http://www.robotis.us/turtlebot-3/}, which is a low-cost, ROS-oriented (Robot Operating System) solution. An accurate model is also provided for Gazebo simulations. The Turtelbot3 Burger model we use is equipped with a Raspberry Pi 3 B+.
Concerning the Ultra-wideband hardware, we use a TREK1000 evaluation kit by Decawave to provide the agent with the localization data that in simulation are obtained via odometry and magnetometer measurements. Fig. \ref{fig:robot} shows the complete robotic platform used during the experimentation.

\begin{table}
\centering
\begin{tabular}{cc}
\hline 
\multicolumn{2}{c}{\textbf{Hyper-parameters}} \\ 
\hline
starting epsilon & $1$ \\
\hline
minimum epsilon & $0.05$ \\
\hline
epsilon decay & $0.998$ \\
\hline
learning rate & $0.00025$\\
\hline
discount factor & $0.99$\\
\hline
sample size & $64$\\
\hline
batch size & $64$\\
\hline
target network update & $2000$\\
\hline
deque memory maxlen & $1000000$\\
\hline 
\end{tabular} 
\caption[Hyperparameters]{Adopted hyperparameters in simulation during the point-to-point agent training.}
\label{Tab:hyperparameters}
\end{table}

\begin{table}[t]
\centering
\begin{tabular}{cc}
%
%
\hline 
\multicolumn{2}{c}{\textbf{Robot settings}} \\ 
\hline
lidar points & $60$\\
\hline
ctrl period & $0.33$\\
\hline
maximum angular speed & $1 rad/s$\\
\hline
maximum linear speed & $0.2 m/s$\\
%
%
\hline
\multicolumn{2}{c}{\textbf{Simulation settings}} \\ 
\hline
time step & $0.0035 s$\\
\hline
max update rate & $2000 s^-1$\\
\hline
timeout & $250 s$ (in sim. time)\\
\hline
\end{tabular} 
\caption[Simulation and robot settings]{Selected settings of the robot and of the simulated environment during the training of the deep reinforcement learning agent.}
\label{Tab:simulation_and_robot_settings}
\end{table}

\subsection{RL Agent Training}
 The training is performed simulating both agent and environment on Gazebo. The robot is controlled by the actor network presented in the methodology. The training is performed in episodes, that means the robot is re-spawned in the same starting point. The objective for each episode is to reach a randomly spawned goal and the reward that is given to the agent depends on it. We use the following equation to provide reward values:

\begin{equation}
    R =  \begin{cases}
    +1000,					& \text{if goal is reached} \\ 
    -200, 					& \text{if collision occurs} \\
    3 \cdot h_R \cdot 10 \cdot |\Delta d| , & \text{else}, \\
\end{cases} 
\end{equation}

\noindent where $\Delta d$ is the difference between distance at current and previous instants of time, and:

\begin{equation}
    h_R= -\left(\omega_{t-1} \cdot \frac{1}{1.2 \cdot	f} - heading\right)^2 + 1
\end{equation}

\begin{figure*}
\centering
\begin{subfigure}{.3\textwidth}
  \centering
  \includegraphics[width=\linewidth]{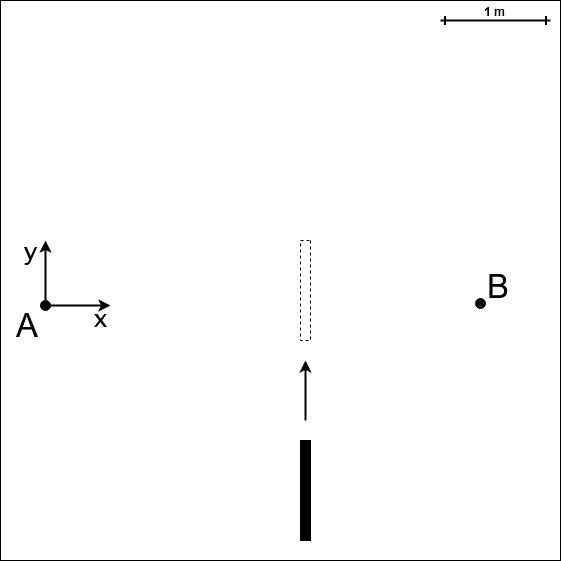}
  \caption{Path: $AB$}
\end{subfigure}
\begin{subfigure}{.3\textwidth}
  \centering
  \includegraphics[width=\linewidth]{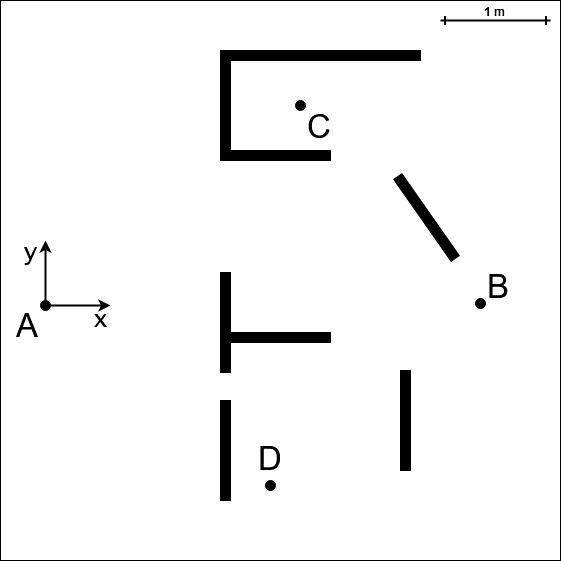}
  \caption{Path: $ABCD$}
\end{subfigure}
\begin{subfigure}{.3\textwidth}
  \centering
  \includegraphics[width=\linewidth]{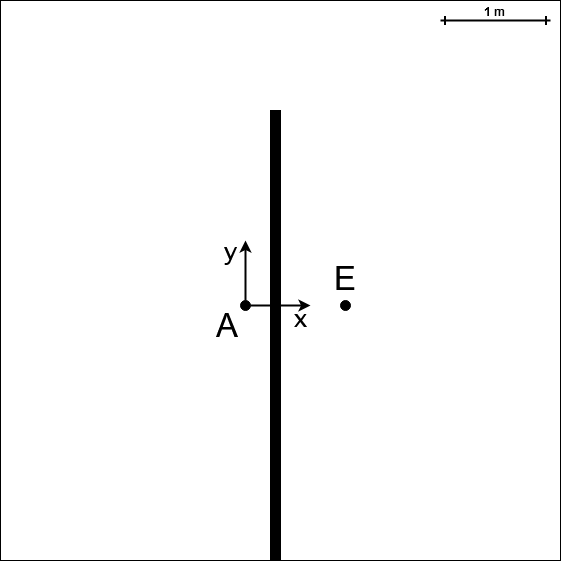}
  \caption{Path: $AE$}
\end{subfigure}
\caption{Scenarios used in the first set of tests for the local planner evaluation. Points positions: $A (0,0)$ \SI{}{\metre}, $B (4.35,0.02)$ \SI{}{\metre}, $C (2.55,2)$ \SI{}{\metre}, $D (2.25,-1.8)$ \SI{}{\metre}, $E (1,0)$ \SI{}{\metre}.}
\label{fig:DRLvDWA}
\end{figure*}

\begin{figure}
    \centerline{\includegraphics[width=0.9\linewidth]{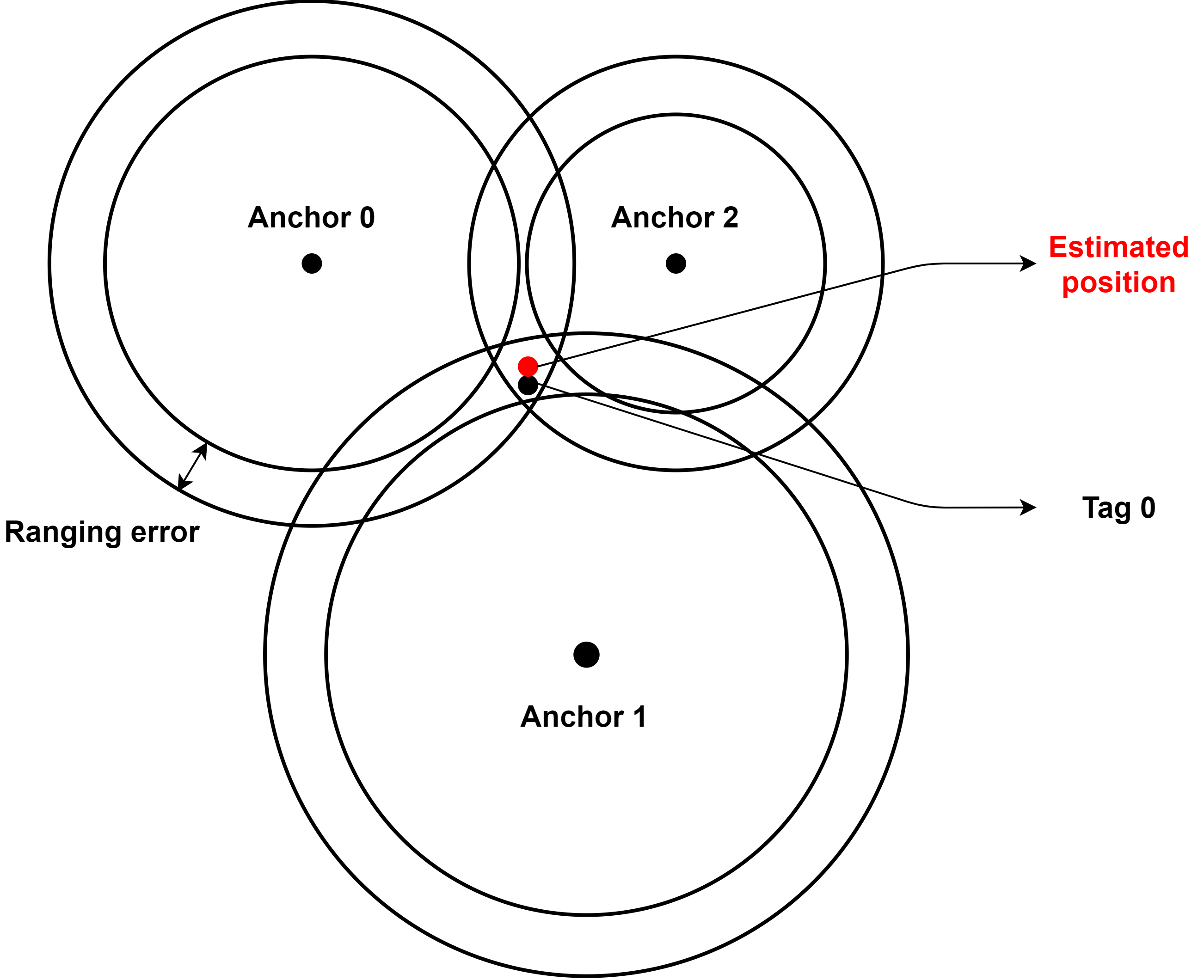}}
    \caption{Estimation of the position in the x-y plan using error affected ranging measurements.}
    \label{fig:UWB Localization}
\end{figure}

\noindent is the heading reward. $\omega$ is the angular speed, while $f$ is the control frequency. The third part of the equation gives a positive reward when the robot is getting closer to the goal ($|\Delta d|$ contribute). Moreover, this value is higher if it is directly pointing it ($ h_R$ contribute). The values of the hyperparameters used in the training phase are shown in Tab. \ref{Tab:hyperparameters}. The \textit{target network update} sets how often the target networks are hardly updated, in terms of steps. The same value of learning rate is used for both actor and critic, equal to 0.00025. The discount factor is set to a value of 0.99, as for the epsilon decay. Tab. \ref{Tab:simulation_and_robot_settings} presents the environment and the robot settings used in simulation.

\subsection{Ultra-wideband settings}

In our experimental setup, we use UWB as the only positioning method, do its robustness against the noise in the localization measurements. The real-time locating system (RTLS) is composed of 5 Decawave EVB1000 boards: 4 placed in fixed positions (anchors) at the corners of the experimental area and one mounted on the robotic platform. The EVB1000 boards are set to communicate using channel 2 (central frequency \SI{3.993}{\giga\hertz}) with a data rate of 6.8 Mbps, a preamble length of 128 symbols and a positioning update rate of 10 Hz. We mount the four anchors on four tripods at slightly different heights, with maximum height set at less than 2 meters. The position of the anchors along the vertical axis strongly affects the precision of the localization in the horizontal (x-y) plan, increasing the height is possible to achieve better performances. Moreover, the robot (target) is able to move outside the area defined by the fixed devices, and this is a critical situation for the computation of the position. The raw ranging data are smoothed using a simple linear Kalman filter. The measurement noise covariance, computed in previous experiments, is set to $\sigma_m^2= 6.67*10^{-4}$ and the process noise covariance ($\sigma_p^2=10^{-4}$) is chosen to obtain the desired behavior from the filter. Finally, the position of the robot is computed as the intersection of the four spheres centered in the anchors' positions with a radius equal to the corresponding ranging measurements, as schematically shown in Fig. \ref{fig:UWB Localization}. This is a typical nonlinear estimation problem that we solve using the Gauss-Newton nonlinear least-squares method, which is a well-suited algorithm for range-based position estimation, as discussed in \cite{yan08feasibility}. At each sampling step, the new position is estimated starting the iteration from the last known point.

\begin{table*}
\centering
\begin{tabular}{l|c|c|c|c|c|}
Scenario & Algorithm & Success rate & $t^\text{mean}$ [\SI{}{\second}]
& $\dot{v}^\text{RMS}$ [\SI{}{\metre\per\square\second}] & $\dot{\omega}^\text{RMS}$ [\SI{}{\radian\per\square\second}] \\ \hline
\multirow{2}{*}{S1}             & DWA       & 1       &  37    & 0.2277 &1.1371 \\
                                & RL+UWB    & 1       &  33    & 0.1342 &1.6382 \\ \hline
\multirow{2}{*}{S2$_{AB}$}      & DWA       & 0.80    &  48    & 0.3016 &1.0022 \\
                                & RL+UWB    & 1       &  45    & 0.1535 &2.6922 \\ \hline
\multirow{2}{*}{S2$_{BC}$}      & DWA       & 0.70    &  97    & 0.2866 &1.6434 \\
                                & RL+UWB    & 0.91    &  65    & 0.1149 &1.4719 \\ \hline
\multirow{2}{*}{S2$_{CD}$}      & DWA       & 0.50    & 129    & 0.2757 &0.9180 \\
                                & RL+UWB    & 0.91    &  94    & 0.1050 &1.5483 \\ \hline
\multirow{2}{*}{S2$_{ABCD}$}    & DWA       & 0.50    & 261    & 0.2880 &1.1879 \\
                                & RL+UWB    & 0.91    & 223    & 0.1225 &1.7528 \\ \hline
\multirow{2}{*}{S3}             & DWA       & 1       &  48    & 0.1920 &1.1390 \\
                                & RL+UWB    & 1       &  31    & 0.1047 &1.4132 \\ \hline 
\end{tabular}
\caption{Experimental results of the first set of tests: comparison with DWA \cite{fox1997dynamic} local planner.}
\label{tab:DRLvsDWA}
\end{table*}

\subsection{Experimental settings}
To prove the robustness and the reliability of the proposed system, we perform several experimentations that can be grouped into two test sets. In the first, we compare our system with a classical one based on the well-known Dynamic Window Approach (DWA) \cite{fox1997dynamic} in different scenarios to prove that our local planner achieves better performances, with lower computational effort. In the second set of tests, we focus on the robustness of the system to UWB localization noise, and we compare it to the performance obtained by humans put in the same testing conditions of the RL agent. The achieved results thoroughly show how the proposed system can represent a reliable and efficient local planner to enable autonomous navigation in unknown and unstructured environments.

In all the following tests, we fix the reference frame on the initial position of the robot and measure the positions with a Leica AT403 Laser Tracker. The main metric for navigation performance is the success rate. Each experiment is considered successful if the robot is able to get within 20 cm to the target position without getting stuck. Since the robot can theoretically reach the goal also with random wandering, we consider a maximum time $t_\text{max}$. If the robot is unable to reach the target position within this time interval, the test is considered failed. Considering the maximum linear speed of \SI{0.22}{\meter\per\second} of the Turtlebot3 Burger and an average path length of \SI{5.5}{\meter} over all the experiments, we consider \SI{180}{\second} as a reasonable value for $t_\text{max}$. Moreover, we consider the mean total time $t^\text{mean}$ as a metric to understand how well the local planner is able to find an optimal solution to the navigation problem and RMS accelerations $\dot{v}^\text{RMS}$, $\dot{\omega}^\text{RMS}$ 
as metrics for navigation smoothness. Finally, collisions with static or moving obstacles are registered for each test, since the ability to avoid them assumes a vital relevance in robotic autonomous navigation.

\subsection{Local Planner Quantitative Evaluation}
The first set of tests is aimed at comparing the proposed local planner with the most used Dynamic Window Approach (DWA) \cite{fox1997dynamic}. We use the ROS implementation of this navigation algorithm, based on the work of Brock et al. \cite{brock1999high}. The two algorithms are compared with repeated tests in three different scenarios:
\begin{enumerate}
    \item the robot has to navigate to the target point autonomously and is suddenly interrupted by a moving obstacle;
    \item the robot has to navigate to three waypoints in a certain order inside a fairly complex environment;
    \item the robot has to reach a goal located behind a wall, with single opening quite far from the goal.
\end{enumerate}

\begin{figure*}[!h]
\centering
\begin{subfigure}{.24\textwidth}
  \centering
  \includegraphics[width=\linewidth]{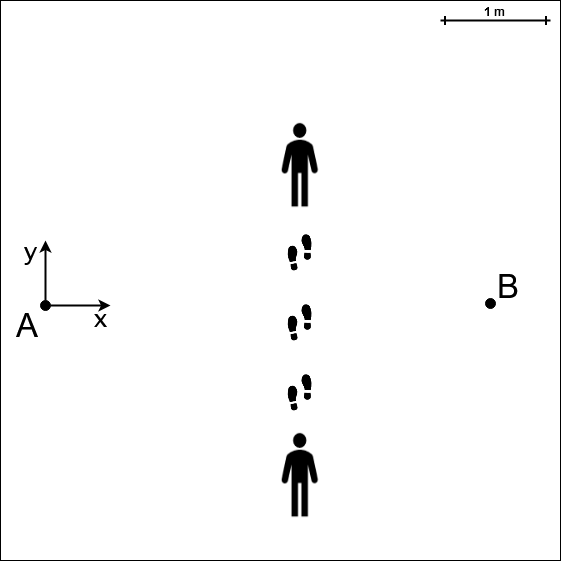}  
  \caption{Path: $AB$}
\end{subfigure}
\begin{subfigure}{.24\textwidth}
  \centering
  \includegraphics[width=\linewidth]{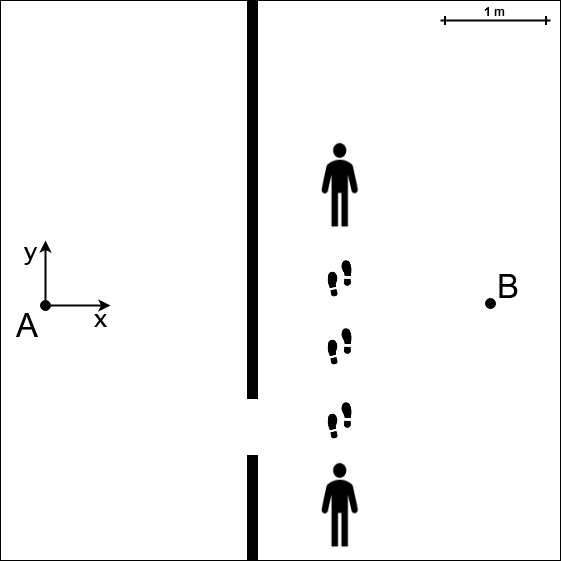}
  \caption{Path: $AB$}
\end{subfigure}
\begin{subfigure}{.24\textwidth}
  \centering
  \includegraphics[width=\linewidth]{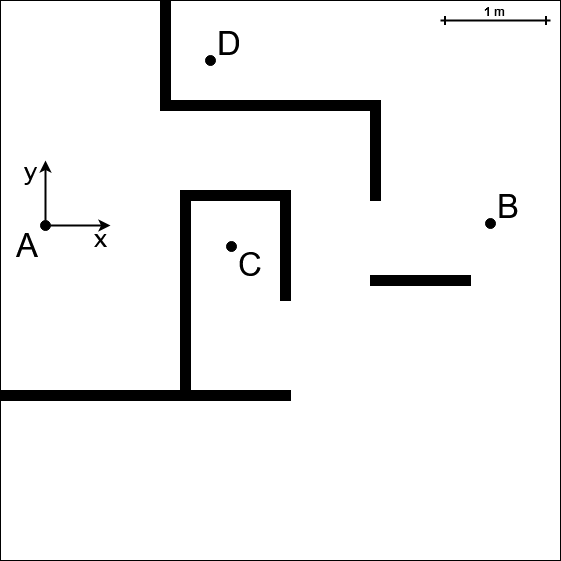}
  \caption{Path: $ABCD$}
\end{subfigure}
\begin{subfigure}{.24\textwidth}
  \centering
  \includegraphics[width=\linewidth]{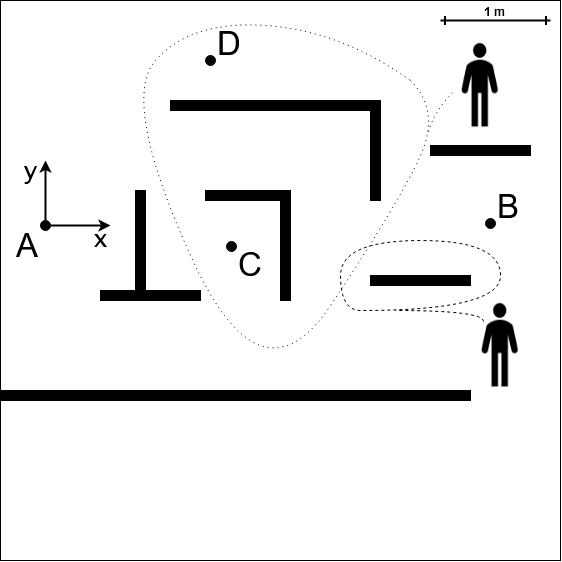}
  \caption{Path: $ACBD$}
\end{subfigure}
\caption{Scenarios used in the second set of tests for the comparison with human control. Points positions: $A (0,0)$ \SI{}{\metre}, $B (4.45,0.02)$ \SI{}{\metre}, $C (1.86,-0.21)$ \SI{}{\metre}, $D (1.65,1.65)$ \SI{}{\metre}.}
\label{fig:DRLvH}
\end{figure*}

Fig. \ref{fig:DRLvDWA} shows a visual presentation of the first tests set scenarios. The first one is particularly useful to evaluate the obstacle avoidance performance of the algorithm and its ability to react to a sudden change in the navigation environment, by following a new safer path to reach the target. In this case, the moving obstacle consists in a panel put in front of the robot while it is navigating towards the target. The second scenario shows the ability to solve subsequent point-to-point tasks in a quite complex and unstructured environment. The robot starts in point $A$ and has to navigate segments $AB$, $BC$, $CD$, subsequently. We evaluate performances both on the single point-to-point tasks and on the whole path $ABCD$. Finally, the last scenario is relevant to judge the ability of the algorithm to adopt local sub-optimal actions that make the robot actually increase the distance from the target, in order to be subsequently able to reach the final goal. In this sense, this kind of situation is useful to evaluate whether the robot is able to escape local minima.

\begin{table*}
\centering
\begin{tabular}{l|c|c|c|c|c|c|}
Scenario & Agent & Success rate & Collisions & $t^\text{mean}$ [\SI{}{\second}]
& $\dot{v}^\text{RMS}$ [\SI{}{\metre\per\square\second}] & $\dot{\omega}^\text{RMS}$ [\SI{}{\radian\per\square\second}] \\ \hline
\multirow{2}{*}{S1}             & Human   & 1       &  0    & 30    & 0.3574 & 2.0783 \\
                                & RL+UWB  & 1       &  0    & 29    & 0.3557 & 3.8413 \\ \hline
\multirow{2}{*}{S2}             & Human   & 1       &  0.25 & 42    & 0.3382 & 2.0012 \\
                                & RL+UWB  & 1       &  0    & 50    & 0.3333 & 4.5058 \\ \hline
\multirow{2}{*}{S3$_{AB}$}      & Human   & 1       &  0.25 & 38    & 1.7703 & 2.1109 \\
                                & RL+UWB  & 1       &  0    & 39    & 0.3513 & 4.1098 \\ \hline
\multirow{2}{*}{S3$_{BC}$}      & Human   & 1       &  0.50 & 36    & 0.3643 & 2.0638 \\
                                & RL+UWB  & 1       &  0    & 29    & 0.3495 & 4.1354 \\ \hline
\multirow{2}{*}{S3$_{CD}$}      & Human   & 0.75    &  0.25 & 97    & 0.3691 & 2.1878 \\
                                & RL+UWB  & 0       &  0    & -     & -      & -      \\ \hline
\multirow{2}{*}{S3$_{ABCD}$}    & Human   & 0.75    &  1    & 161   & 0.3696 & 2.1490 \\
                                & RL+UWB  & 0       &  0    & -     & -      & -      \\ \hline
\multirow{2}{*}{S4$_{AC}$}      & Human   & 1       &  0    & 49    & 0.3227 & 2.1171 \\
                                & RL+UWB  & 1       &  0    & 49    & 0.3230 & 4.5523 \\ \hline
\multirow{2}{*}{S4$_{CB}$}      & Human   & 1       &  0.25 & 40    & 0.3224 & 2.3882 \\
                                & RL+UWB  & 1       &  0    & 28    & 0.3393 & 4.4387 \\ \hline
\multirow{2}{*}{S4$_{BD}$}      & Human   & 1       &  0    & 49    & 0.3420 & 2.1848 \\
                                & RL+UWB  & 1       &  0    & 25    & 0.3280 & 4.3618 \\ \hline
\multirow{2}{*}{S4$_{ACBD}$}    & Human   & 1       &  0.25 & 137   & 0.3290 & 2.2301 \\
                                & RL+UWB  & 1       &  0    & 102   & 0.3301 & 4.4509 \\ \hline
\end{tabular}
\caption{Experimental results of the second set of tests: comparison with human control.}
\label{tab:DRLvsH}
\end{table*}

We perform a total of 30 tests for both the algorithms in the three different scenarios. Tab. \ref{tab:DRLvsDWA} summarizes the experimentation. In general, our approach has a higher success rate and requires, on average, less time to reach the target. It gets lower linear accelerations, but higher angular ones, resulting in a lower smoothness on the angular control. The second scenario appears to be the toughest one, in particular in its third task $CD$, where the DWA success rate drops to 0.5. In all these tests, we register no collisions with both the algorithms. However, the main advantage of the proposed local planner is its computational effort. We achieve up to \SI{400}{\hertz} control frequency the proposed RL planner. On the other hand, since the DWA is an optimization algorithm, on the same machine it ranges between \SI{0.5}{\hertz} and \SI{5}{\hertz}. This dramatic improvement in computational efficiency allows for the proposed local planner to be completely run in an embedded system on the robot itself, without the need of a powerful machine as classic algorithms as DWA do. We deploy the RL agent on a Raspberry Pi3 B+ embedded computer, and we are able to achieve a real-time control at about \SI{30}{Hz}.

\subsection{Noise Robustness and Human Comparison}
\begin{figure}
    \centering
    \includegraphics[width=0.8\linewidth]{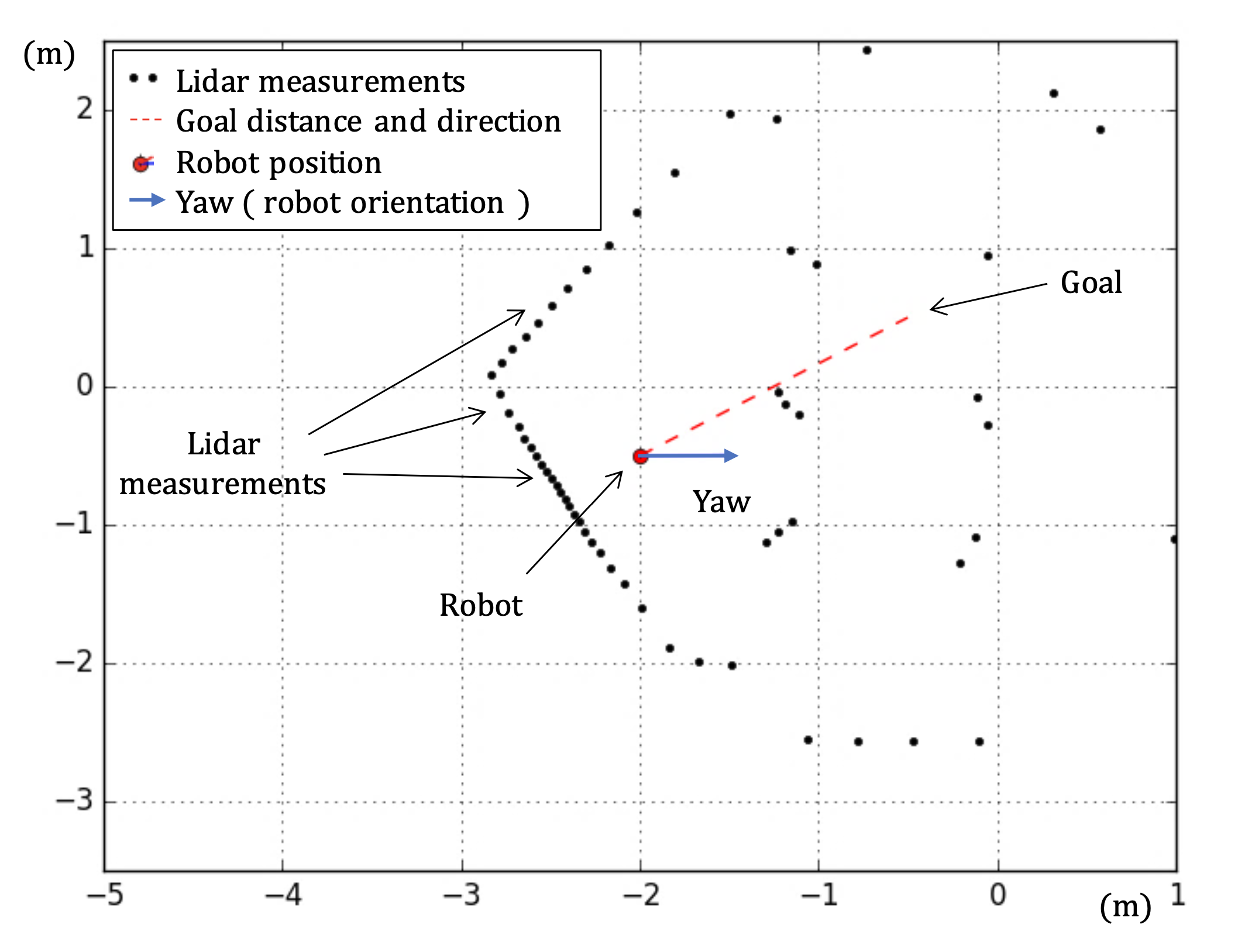}
    \caption{Human interface during the second tests set. Testers are allowed to see lidar measurements, robot pose and goal distance and direction.}
    \label{fig:humaninterface}
\end{figure}

The second set of tests is aimed at comparing the proposed algorithm with human performance, as well as demonstrate how the RL+UWB system is highly robust against localization noise. In literature, RL agents are frequently compared to human agents to prove their control performance in complex tasks \cite{mnih2015human,silver2016mastering,mirowski2016learning,silver2018general}. We perform such comparison by putting several people in the same experimental conditions of the RL agent. Human testers are kept in a different room with respect to the experimental environment and can see in real-time the robot position, the goal and the 1-D lidar range measurements, that are exactly the same information available to the RL planner. Fig. \ref{fig:humaninterface} presents the interface shown to human testers during the experimentation. Both humans and RL agent are tested in the following scenarios, shown in Fig. \ref{fig:DRLvH}:
\begin{enumerate}
    \item the robot has to navigate to the target point autonomously and is suddenly interrupted by a person;
    \item the robot has to pass through a small opening partially occluded by a moving obstacle; 
    \item the robot has to navigate to three waypoints in a certain order inside a fairly complex environment;
    \item the robot has to navigate to three waypoints in a certain order inside a fairly complex environment with both static and moving obstacles (people wandering in the scenario).
\end{enumerate}

\begin{figure}
\centering
\includegraphics[width=1\linewidth]{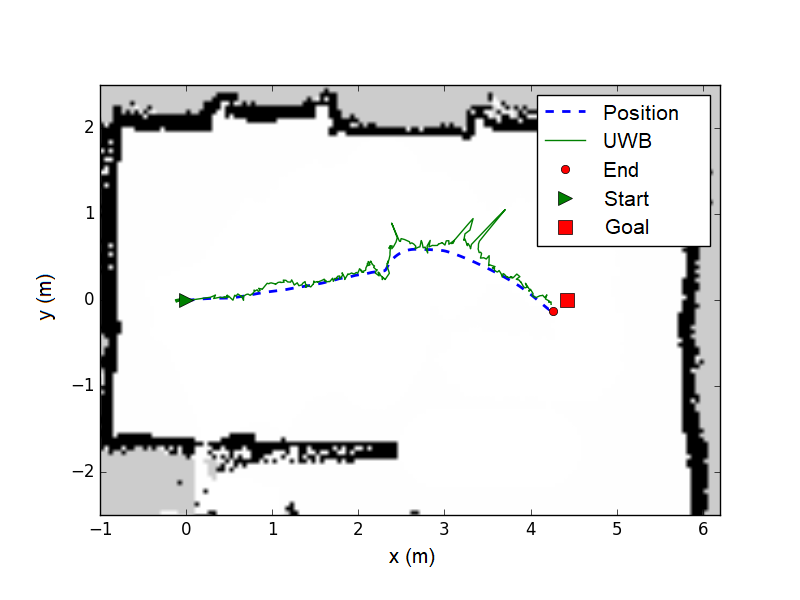}  
\caption{Trajectory followed in a test in the first scenario (path $AB$, interrupted by a sudden moving person). The UWB added noise is clearly visible and shows how the proposed local planner is highly robust against positioning errors.}
\label{fig:UWBnoise}
\end{figure}

In all these tests, Gaussian noise is superimposed to UWB measurements in order to evaluate the robustness of the system to localization errors. Higher uncertainty in the UWB positioning is also caused by the presence of people in the environments (scenarios 1, 2, 4) who obstruct the anchors and cause the NLOS (non-line of sight) condition. Fig. \ref{fig:UWBnoise} shows an example of the trajectory followed by the robot in the first scenario (path $AB$, interrupted by a sudden moving person). The noisy signal of the UWB clearly gives a high uncertainty on the position of the robot. However, the RL local planner is highly robust against localization errors and it is able to reach the goal.


Tab. \ref{tab:DRLvsH} presents the results of the second set of tests. In general, the RL+UWB appears to have similar performances to human control, even in the presence of localization noise. The proposed algorithm appears to be particularly able in avoiding obstacles, while humans result more subject to collisions in complex environments. One interesting thing to notice is that the RL+UWB is completely unable to solve the $CD$ task of scenario 3, when the robot is surrounded by walls on three edges. It remains stuck, repeating the same actions over and over. This behavior can be explained by the absence of memory in this kind of planners, that makes them unable to escape from too narrow local minima. Humans are able to analyze subsequent states and can understand how the environment is actually disposed, while the RL agent simply reacts to the current state and cannot build an environment map.


\section{Conclusion}
In this paper, we proposed a novel indoor local motion planner based on a strict synergy between an autonomous agent trained with deep reinforcement learning and ultra-wideband localization technology. Indoor autonomous navigation is a challenging task, and localization techniques can generate noisy and unreliable signals. Moreover, due to the high complexity of typical environments, hand-tuned classical methodologies are highly prone to failure and require access to a large number of computational resources. The extensive experimentation and evaluations of our research proved that our low-cost and power-efficient solution has comparable performance with classical methodologies and is robust to noise and scalable to dynamic and unstructured environments.


\section*{acknowledgments}
This work has been developed with the contribution of the Politecnico di Torino Interdepartmental Centre for Service Robotics PIC4SeR (https://pic4ser.polito.it) and SmartData@Polito (\url{https://smartdata.polito.it}).


\bibliographystyle{IEEEtran}
\bibliography{mainArXiv}

\vfill\break

\begin{IEEEbiography}[{\includegraphics[width=1in,height=1.25in,clip,keepaspectratio]{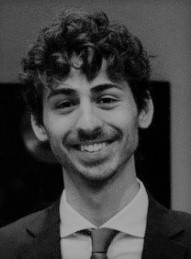}}]{Enrico Sutera} is a research fellow at PIC4SeR (\url{https://pic4ser.polito.it/}). He received a Bachelor's Degree in Mechanical Engineering in 2017 and a Master's Degree in Mechatronics Engineering in 2019, with the thesis "Deep Reinforcement Learning and Ultra-Wideband for autonomous navigation in service robotic applications". He is currently working on indoor robotics along with deep reinforcement learning.
\end{IEEEbiography}

\begin{IEEEbiography}[{\includegraphics[width=1in,height=1.25in,clip,keepaspectratio]{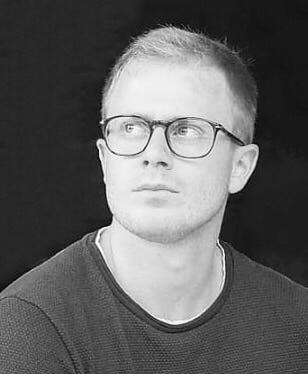}}]{Vittorio Mazzia} is a Ph.D. student in Electrical, Electronics and Communications Engineering working with the two Interdepartmental Centres PIC4SeR (\url{https://pic4ser.polito.it/}) and SmartData (\url{https://smartdata.polito.it/}). He received a master's degree in Mechatronics Engineering from the Politecnico di Torino, presenting a thesis entitled "Use of deep learning for automatic low-cost detection of cracks in tunnels," developed in collaboration with the California State University. His current research interests involve deep learning applied to different tasks of computer vision, autonomous navigation for service robotics, and reinforcement learning. Moreover, making use of neural compute devices (like Jetson Xavier, Jetson Nano, Movidius Neural Stick) for hardware acceleration, he is currently working on machine learning algorithms and their embedded implementation for AI at the edge. 
\end{IEEEbiography}

\begin{IEEEbiography}[{\includegraphics[width=1in,height=1.25in,clip,keepaspectratio]{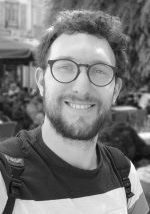}}]{Francesco Salvetti} is currently a Ph.D. student in Electrical, Electronics and Communications Engineering in collaboration with the two interdepartmental centers PIC4SeR (\url{https://pic4ser.polito.it/}) and Smart Data (\url{https://smartdata.polito.it/}) at Politecnico di Torino, Italy. He received his Bachelor's Degree in Electronic Engineering§ in 2017 and his Master’s Degree in Mechatronics Engineering in 2019 at Politecnico di Torino. He is currently working on Machine Learning applied to Computer Vision and Image Processing in robotics applications.
\end{IEEEbiography}

\begin{IEEEbiography}[{\includegraphics[width=1in,height=1.25in,clip,keepaspectratio]{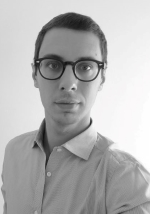}}]{Giovanni Fantin} s a research fellow at PIC4SeR (\url{https://pic4ser.polito.it/}). In 2019, he achieved the Master's Degree in Mechatronics Engineering at Politecnico di Torino discussing the thesis "UWB localization system for partially GPS denied robotic applications". He is currently working on a PRIN (progetto di rilevante interesse nazionale) about new generation ultra-wideband technologies with a particular focus on multi-robot cooperation to perform localization.
\end{IEEEbiography}

\begin{IEEEbiography}[{\includegraphics[width=1in,height=1.25in,clip,keepaspectratio]{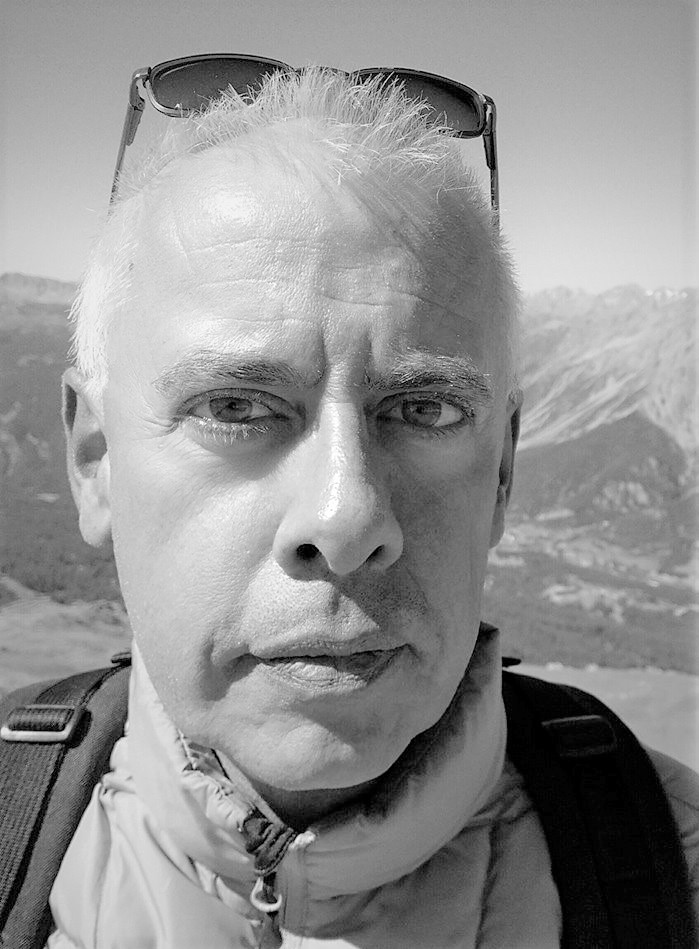}}]{Marcello Chiaberge} is currently Associate Professor within the Department of Electronics and Telecommunications, Politecnico di Torino, Turin, Italy. He is also the Co-Director of the Mechatronics Lab, Politecnico di Torino
(\url{www.lim.polito.it}), Turin, and the Director and the Principal Investigator of the new Centre for Service Robotics (PIC4SeR, \url{https://pic4ser.polito.it/}), Turin. He has authored more than 100 articles accepted in international conferences and journals,
and he is the coauthor of nine international patents. His research interests include
hardware implementation of neural networks and fuzzy systems and the design and implementation of reconfigurable real-time computing architectures. \end{IEEEbiography}
\vfill

\end{document}